\documentclass[runningheads]{llncs}

\usepackage{iciap}

\usepackage{graphicx}
\usepackage{booktabs}
\usepackage{siunitx}
\usepackage{multirow}
\usepackage{adjustbox}
\usepackage{algorithm}
\usepackage{algorithmic}
\usepackage[accsupp]{axessibility}  

\usepackage[pagebackref,breaklinks,colorlinks,citecolor=iciapblue]{hyperref}

\usepackage{orcidlink}

\title{Beyond Linear Bottlenecks: Spline-Based Knowledge Distillation for Culturally Diverse Art Style Classification} 

\titlerunning{Beyond Linear Bottlenecks: Spline-Based Knowledge Distillation}

\author{Abdellah Zakaria Sellam\inst{1,2}\orcidID{0009-0003-6876-2220} \and
Salah Eddine Bekhouche\inst{3}\orcidID{0000-0001-5538-7407} \and
Cosimo Distante\inst{1}\orcidID{0000-0002-1073-2390} \and
Abdelmalik Taleb-Ahmed\inst{4}\orcidID{0000-0001-7218-3799}}

\authorrunning{AZ. Sellam et al.}

\institute{Institute of Applied Sciences and Intelligent Systems – CNR, Via per Monteroni, 73100 Lecce, Italy \and
Department of Innovation Engineering, University of Salento \and
UPV/EHU, University of the Basque Country, 20018 San Sebastian, Spain \and
Université Polytechnique Hauts-de-France, Université de Lille, CNRS, 59313 Valenciennes, France
}

\begin{document}
\maketitle
\begin{abstract}
Art style classification remains a formidable challenge in computational aesthetics due to the scarcity of expertly labeled datasets and the intricate, often nonlinear interplay of stylistic elements. While recent dual-teacher self-supervised frameworks reduce reliance on labeled data, their linear projection layers and localized focus struggle to model global compositional context and complex style-feature interactions. We enhance the dual-teacher knowledge distillation framework to address these limitations by replacing conventional MLP projection and prediction heads with Kolmogorov–Arnold Networks (KANs). Our approach retains complementary guidance from two teacher networks, one emphasizing localized texture and brushstroke patterns, the other capturing broader stylistic hierarchies while leveraging KANs’ spline-based activations to model nonlinear feature correlations with mathematical precision. Experiments on WikiArt and Pandora18k demonstrate that our approach outperforms the base dual-teacher architecture in Top-1 accuracy. Our findings highlight the importance of KANs in disentangling complex style manifolds, leading to better linear probe accuracy than MLP projections.
\keywords{KAN \and Dual-teacher \and Classification \and knowledge-distillation \and Projection}%
\end{abstract}

\section{Introduction}
\label{sec:introduction} 

Art style classification is vital in computational aesthetics, bridging human artistic expression and machine interpretation. It enables algorithms to analyze and categorize various artistic styles while recognizing patterns in texture, composition, and color influenced by cultural contexts.

Despite the success of supervised deep learning methods~\cite{johnson2017neural}, their dependence on large labeled datasets poses challenges in the art domain, where expert annotations are limited and costly. Self-Supervised Learning (SSL) offers promising alternatives by utilizing unlabeled data, but existing SSL frameworks in art analysis still face hurdles.

Earlier methods relied on hand-crafted features and traditional classifiers like Support Vector Machines (SVM) and k-Nearest Neighbors (kNN)~\cite{karayev2013stylearxiv}. These approaches, while interpretable, often fail to generalize and capture abstract stylistic patterns~\cite{falomir2018qartlearn}. The rise of deep learning has advanced the field~\cite{zhou2024three}, with Convolutional Neural Networks (CNNs) outperforming traditional methods by automatically learning hierarchical visual representations~\cite{peterson2018learninghierarchicalvisualrepresentations}. Transfer learning has further improved performance; Cetinic et al.~\cite{cetinic2018fine} demonstrated that networks pre-trained on scene recognition tasks perform better in art classification than object-centric models. However, these supervised approaches remain limited by their reliance on labeled data, which is scarce in curated art datasets.

Recent research has explored unsupervised and self-supervised learning to tackle challenges in image processing. Triplet networks using Gram matrices have shown promise in weakly supervised style grouping \cite{gairola2020unsupervised}, while multi-task self-supervised frameworks have incorporated aesthetic elements like brightness and composition \cite{zhang2022threeelement}. However, these approaches can increase computational complexity and necessitate careful hyperparameter tuning. Although frameworks such as MoCo \cite{he2020moco} and SimCLR \cite{chen2020simclr} are effective for general image tasks, they often struggle to isolate style-specific features vital for recognizing subtle artistic traits.

Previous research in SSL-driven art classification mainly concentrated on multi-teacher architectures and handcrafted feature alignment. For instance, Luo et al.~\cite{luo2025} introduce a dual-teacher framework that uses Gram matrices and relation alignment losses to distill style features. However, it increases training complexity and limits adaptability across artistic domains. Additionally, traditional multilayer perceptron (MLP) projection heads struggle with the complex nonlinearities of artistic styles due to fixed activation functions.

We introduce a novel self-supervised framework for art-style classification by embedding \textbf{Kolmogorov–Arnold Networks (KANs)} as nonlinear projection heads in place of conventional MLPs~\cite{liu2024kan}.  Grounded in the Kolmogorov–Arnold representation theorem, each edge in a KAN is parameterized by a \emph{learnable spline-based univariate function} rather than a fixed activation. This design adaptively sculpts basis functions to capture subtle, high-order interactions among style features, enabling the precise disentanglement of overlapping artistic attributes.
Our contributions are twofold:

\begin{itemize}
  \item \emph{Integration of KANs into SSL for Art Style.} We are the first to use KAN-based projection heads in a contrastive self-supervised learning pipeline for art-style classification, showing significant improvements over standard MLP heads.
  \item \emph{Enhanced Modeling of Fine-Grained Artistic Variations.} Our approach uses KANs to capture nuanced style differences often missed by traditional architectures, enhancing performance on fine-grained classification tasks.
\end{itemize}

This paper is structured as follows: Section~\ref{Related_works} overviews the foundational work on semi-supervised learning (SSL) and art style classification. In Section~\ref{sec:methods}, we describe our methodology, highlighting the architectural advantages of KANs and their synergy with gram matrix alignment. 
Section~\ref{sec:experiments} presents the experimental validation of our approach, providing detailed results and metrics. Following this, Section~\ref{sec:Discussion} offers an in-depth analysis of the findings, discussing their implications, limitations, and potential future research directions. Finally, Section~\ref{sec:conclusion} summarizes our key findings and contributions to the field.

\section{Related works}
\label{Related_works}

The evolution of art style classification has been influenced by feature engineering, deep learning, and domain adaptation. Early methods, such as \cite{karayev2013recognizing}, depended on handcrafted features like color histograms and texture descriptors, paired with shallow classifiers like SVM and kNN. This approach, while interpretable, struggled with generalization across diverse artistic styles.

The introduction of deep learning brought a significant shift. \cite{krizhevsky2012imagenet} showcased the power of convolutional neural networks (CNNs) for hierarchical feature extraction. However, adapting these methods to art was challenging due to domain differences. \cite{simonyan2014very} developed deeper architectures focused on object-centric tasks to improve feature abstraction.
Transfer learning is crucial, as noted by \cite{cetinic2019}, with fine-tuned models for scene recognition surpassing those pre-trained on ImageNet. Hierarchical classification and disentangled representations \cite{elgammal2018} refine style categories but heavily depend on labeled datasets. To lessen annotation dependence, \cite{zhang2023multitask} proposed a multi-task self-supervised framework using compositional rules. \cite{ressl2023relation} established relational consistency via Gram matrix alignment, and \cite{kan2024nonlinear} introduced Kolmogorov-Arnold Networks (KANs) for nonlinear style-feature interactions. Self-supervised learning (SSL) is key for robust representation learning from unlabeled art datasets. Early methods like MoCo \cite{he2020} and SimCLR \cite{chen2020} faced batch size issues, which BYOL addressed by removing negative samples and ensuring encoder output consistency. \cite{garg2022selflabelingrefinementrobustrepresentation} further refined this with SimSiam, using stop-gradient operations to prevent collapse, though challenges in fine-grained style discrimination remain. 

Recent innovations have focused on \textbf{style-aware SSL}. Building on \cite{ressl2023relation}, our framework integrates Gram matrices to enforce relational consistency across augmented views. Concurrently, \cite{liu2024} proposed \textit{ArtCLR}, a domain-specific SSL variant incorporating art-historical metadata as weak supervision.  
Knowledge distillation (KD), first introduced by \cite{hinton2015distilling}, has evolved into a versatile tool for compressing models and transferring domain-specific knowledge. Traditional single-teacher distillation \cite{fukuda2017} suffered from "knowledge bottleneck" issues. Multi-teacher frameworks addressed this by aggregating complementary expertise: \cite{pham2020distillation} fused logits from multiple teachers using temperature-weighted ensembling.  
Multi-teacher strategies~\cite{fukuda2017efficient, pham2023collaborative}, enhanced student models through diverse supervision.
Luo et al.~\cite{luo2025dual} introduced a dual-teacher contrastive framework using Gram matrices and relation alignment loss. However, their linear MLP projection network limits non-linear feature interactions, a critical constraint for modeling complex artistic styles.
We address this limitation by replacing the linear MLP with a \textbf{Kolmogorov-Arnold Network (KAN)}~\cite{liu2024kan}. This architecture preserves Luo et al.~\cite{luo2025dual} dual-teacher distillation and Gram matrix integration while enabling exponential efficiency in learning stylistic manifolds. KAN's spline-based design mitigates spectral bias, enhancing the disentanglement of subtle style features and a marked improvement over rigid MLP parameterizations.

\section{Methods}
\label{sec:methods}

\subsection{Architecture Overview}
\begin{figure}[t]
    \centering
    \includegraphics[width=0.9\linewidth]{"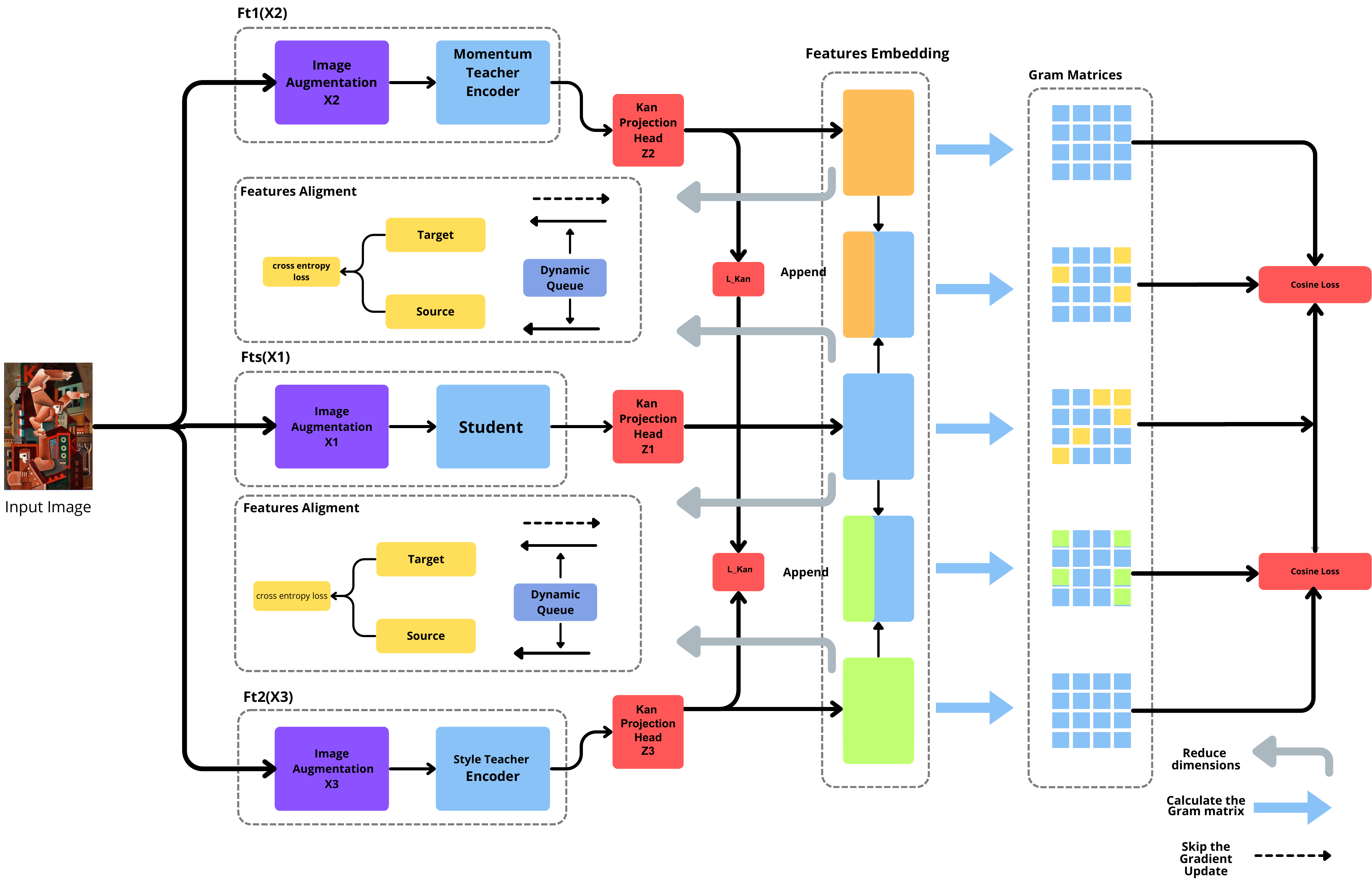"}
    \caption{Overall architecture of our proposed network. For each input image, three augmented views X1, X2, and X3 are generated. X1 and X3 are fed to the two teacher networks,
respectively. And X2 is input to the student network. Two teachers collaboratively guide the student model, to ensure alignment with the teachers’ guidance.}
    \label{fig:arch}
\end{figure}
The proposed architecture, illustrated in Figure \ref{fig:arch}, integrates a dual-teacher knowledge distillation framework with a Kolmogorov–Arnold Network (KAN) projection head and a style-aware feature alignment to address the challenges of art style classification. At its core, the system comprises three main pathways: two teachers and a student network, all trained collaboratively under a self-supervised paradigm.
 The training pipeline begins by generating three augmented views per input image: a weakly augmented sample x1 for the Momentum Teacher, a strongly augmented sample x2
for the Student, and another weakly augmented view x3 for the Style Teacher. 
This asymmetric augmentation strategy creates a controlled discrepancy between teacher and student inputs, forcing students to learn invariances to challenging transformations while teachers maintain stable feature extraction.

Each augmented view is passed through a corresponding encoder: the Momentum Teacher ($F_{t1}$), the Style Teacher ($F_{t2}$), and the Student ($F_s$). These encoders are typically based on vision transformers or convolutional backbones and produce feature maps $h_1, h_2, h_3 \in \mathbb{R}^{C \times H' \times W'}$ that capture local and global semantic structures. These encoded features are then passed to a shared projection head, where the key contribution is that all branches use a Kolmogorov-Arnold Network (KAN) as their projection function.

Unlike traditional MLPs that rely on fixed activation functions like ReLU or GELU, KAN uses a two-level composition of learnable univariate B-spline functions. For each feature dimension $h_i^p$, KAN computes:

\begin{equation}
    z_i = \mathrm{KAN}(h_i) = \sum_{q=1}^{2n+1} \Phi_q \left( \sum_{p=1}^n \phi_{q,p}(h_i^p; \theta_{q,p}) \right)
\end{equation}

Here, $\phi_{q,p}$ denotes a univariate cubic B-spline function with learnable control points $\theta_{q,p}$ and adaptive knot spacing. At the same time, $\Phi_q$ are trainable composition weights initialized from a Gaussian distribution and constrained by $\ell_2$-normalization. This construction allows each unit in the projection head to adaptively shape its activation response to suit local curvature, style shifts, or semantic boundaries in the data.

The outputs of these projections, $z_1, z_2, z_3 \in \mathbb{R}^d$, are used in multiple downstream objectives. First, the feature bank is a FIFO queue $Q$ storing previously computed teacher projections, enabling scalable contrastive learning. The Student embedding $z_2$ is compared against both current teacher embeddings and historical entries in $Q$, using cosine similarity to form soft probability distributions:

\begin{equation}
    P_i = \mathrm{softmax}\left( \frac{\cos(z_2, z_i)}{\tau} \right), \quad i=1,3
\end{equation}

At each training step, the Student's embedding $z_2$ is compared to the teacher's embeddings via cosine similarity, generating three probability distributions. These distributions are aligned using KL divergence, yielding the relation alignment loss:

\begin{equation}
    \mathcal{L}_{\text{Relation}} = D_{\text{KL}}(P_1 \parallel P_3) + D_{\text{KL}}(P_2 \parallel P_3)
\end{equation}

Which ensures that the Student mimics the behavior of both teachers across current and historical embedding states.

In parallel, the architecture enforces feature alignment at the style level. The encoder features $h_i$ are used to construct Gram matrices $G_i = \frac{1}{HW}h_i^\top h_i$, capturing the second-order channel correlations (texture/style). These Gram representations are then compared between students and teachers via cosine similarity in Frobenius space, forming the style alignment loss:

\begin{equation}
    \mathcal{L}_{\text{Style}} = 1 - \frac{\langle G_1, G_2 \rangle_F}{\|G_1\|_F \|G_2\|_F} + 1 - \frac{\langle G_3, G_2 \rangle_F}{\|G_3\|_F \|G_2\|_F}
\end{equation}

Which encourages the Student to preserve style-aware structural features.

The KAN projection head is heavily regularized to prevent overfitting and encourage meaningful representations. This includes:
\begin{itemize}
    \item[(1)] an L1 sparsity loss $\mathcal{L}_{L1} = \sum_{q,p} \|\theta_{q,p}\|_1$ on spline parameters,
    \item[(2)] a smoothness loss $\mathcal{L}_{\text{smooth}} = \sum_s \int |f_s''(x)|^2 dx$ that penalizes sharp spline bends, and
    \item[(3)] a segment deactivation loss $\mathcal{L}_{\text{deact}}$ which randomly turns off parts of spline activations during training, similar to dropout but localized to spline segments.
\end{itemize}

These components are combined:

\begin{equation}
    \mathcal{L}_{\text{KAN}} = \lambda_{L1}\mathcal{L}_{L1} + \lambda_{\text{smooth}}\mathcal{L}_{\text{smooth}} + \mathcal{L}_{\text{deact}}
\end{equation}

The final loss combines relation alignment, style preservation, and KAN regularization:

\begin{equation}
    \mathcal{L}_{\text{total}} = \mathcal{L}_{\text{Relation}} + 0.5 \cdot \mathcal{L}_{\text{Style}} + \mathcal{L}_{\text{KAN}}
\end{equation}

The student encoder and KAN head are updated via backpropagation on $\mathcal{L}_{\text{total}}$, while the two teacher branches are updated using Exponential Moving Average (EMA):

\begin{equation}
    \theta_t^{(t)} = m \cdot \theta_t^{(t-1)} + (1-m) \cdot \theta_s^{(t)}
\end{equation}

where $m=0.99$. This strategy ensures slow, stable updates of teacher networks to guide the more volatile students. The architecture fuses contrastive, structural, and style-aware supervision through a KAN-driven, spline-regularized embedding mechanism, making it highly effective for domains like artistic or stylized image classification where rigid features and weak labels fail.

\section{Experiments}
\label{sec:experiments}
This section introduces the Pandora 18K and WikiArt datasets, highlighting their importance for fine-grained artistic style analysis. We present our dual-teacher self-supervised framework that trains a student model using two teachers: one with a traditional MLP projection head and another with our Kolmogorov-Arnold Network (KAN) projection head, optimized with a composite contrastive loss. After training, we freeze the Student’s backbone to preserve learned representations and apply a linear evaluation protocol to assess feature quality in style classification tasks. Finally, we analyze how KAN enhances feature separability for intricate artistic styles and compare the performance across backbone architectures to demonstrate the framework's robustness.
\subsection{Dataset}
Our experiments are conducted on two publicly available datasets, as detailed below. For all datasets, we adopt a data split that mirrors the Dual Teacher paper regarding the number of training and validation samples, but with a different random seed. The precise splits are included in our code repository to ensure reproducibility.

\textbf{WikiArt Dataset.} The WikiArt dataset~\cite{Karayev_2014_BMVC} includes over 80,000 artworks across 25 style categories created by 195 artists. We selected the 10 categories with the most images, resulting in a subset of 53,072 images: 37,146 for training, 7,956 for validation, and 7,970 for testing. This approach ensures ample samples per class for effective feature projection and evaluation of our dual-teacher KAN framework.

\textbf{Pandora18k Dataset.} We use the Pandora18k dataset~\cite{7926652}, which contains 18,038 images from various artistic genres and photographic styles. To ensure consistency with previous work on the Dual Teacher framework, we replicate its train/validation/test proportions using an independent random seed for reproducibility and to prevent data leakage. All split indices and preprocessing scripts are available in our public repository.

\subsection{Implementation}
The network is optimized using Stochastic Gradient Descent (SGD). Hyperparameters, including batch size, initial learning rate, and input resolution, are empirically determined for optimal performance on each dataset. For the \textbf{WikiArt} dataset, we adopt a batch size of 32, an initial learning rate of 0.0075, and resize input images to $480 \times 480$ pixels. These settings balance the preservation of high-resolution artistic details with stable training dynamics. Experiments for the \textbf{Pandora18k} dataset show that a batch size of 16, an initial learning rate of 0.001, and $352 \times 352$ pixel dimensions achieve effective performance while maintaining computational efficiency. 

Momentum coefficients $\alpha$ and $\beta$ are fixed at 0.99. The learning rate follows a linear warm-up schedule followed by cosine annealing, starting from the specified base rates. Training is conducted for 25 epochs on an NVIDIA Quadro 4500 GPU. The KAN projection head employs a $5 \times 5$ transformation grid with cubic spline functions of order 3 to capture high-order nonlinear feature interactions. All implementations are based on PyTorch 1.12.1 with CUDA 12.4 support.

\subsection{Results}

In Table~\ref{tab:performance-multi-dataset-variant}, we systematically evaluate the impact of substituting the original MLP projection head in our dual-teacher self-supervised framework with the proposed KAN module across three leading backbone architectures, EfficientNet-B0, ConvNeXt-Base, and ViT-Base, on the Pandora18k and WikiArt benchmarks. On Pandora18k, integrating KAN into EfficientNet-B0 yields a 0.92 \% increase in Top-1 accuracy, a 1.09 \% gain in Top-5 accuracy, a 1.23 \% improvement in precision, a 0.84 \% rise in recall and a 1.01 \% uplift in F1-score; ConvNeXt-Base with KAN attains a 1.03 \% boost in Top-1 accuracy, a 0.36 \% gain in Top-5 accuracy, a 0.99 \% increase in precision, a 0.89 \% rise in recall and a 0.97 \% enhancement in F1-score; ViT-Base augmented by KAN delivers a 0.39 \% improvement in Top-1 accuracy, a 0.14 \% gain in Top-5 accuracy, a 0.50 \% increase in precision, a 0.40 \% rise in recall and a 0.31 \% uplift in F1-score. On the more challenging WikiArt dataset, EfficientNet-B0+KAN exhibits a marginal 0.03 \% decrease in Top-1 accuracy alongside a 0.37 \% gain in Top-5 accuracy and a 0.42 \% rise in precision, counterbalanced by a 0.72 \% drop in recall and a 0.29 \% reduction in F1-score; ConvNeXt-Base+KAN achieves a 0.87 \% increase in Top-1 accuracy, a 0.46 \% gain in Top-5 accuracy, a 0.63 \% uplift in precision, a 0.93 \% rise in recall and a 0.76 \% improvement in F1-score; ViT-Base+KAN secures a 0.23 \% enhancement in Top-1 accuracy, a 0.33 \% gain in Top-5 accuracy, a 0.96 \% increase in precision, a 0.21 \% rise in recall and a 0.60 \% uplift in F1-score. These results demonstrate that KAN consistently refines feature clustering and mitigates misclassification, particularly in fine-grained, imbalanced settings, thereby substantively strengthening generalization in self-supervised representation learning.

\begin{table}[t]
\centering
\footnotesize
\setlength{\tabcolsep}{3pt}
\renewcommand{\arraystretch}{0.7}
\caption{Performance comparison of models (Base vs.\ KAN) on the \textbf{Pandora18k} and \textbf{WikiArt} datasets.}
\label{tab:performance-multi-dataset-variant}
\begin{adjustbox}{width=\linewidth}
\begin{tabular}{
  l 
  l 
  l 
  S[table-format=2.2] 
  S[table-format=2.2] 
  S[table-format=2.2] 
  S[table-format=2.2] 
  S[table-format=2.2] 
}
\toprule
\textbf{Dataset} & \textbf{Model} & \textbf{Variant} & {\textbf{Top‑1\,(\%)}} & {\textbf{Top‑5\,(\%)}} & {\textbf{Prec.\,(\%)}} & {\textbf{Rec.\,(\%)}} & {\textbf{F1\,(\%)}} \\
\midrule
\multirow{8}{*}{Pandora18k}

 & EfficientNet‑B0 \cite{tan2019efficientnet} & Base & \underline{49.16} & 88.98 & \underline{49.32} & \underline{49.65} & \underline{49.04} \\
 & & KAN  & \textbf{50.08}   & 90.07 & \textbf{50.55}   & \textbf{50.49}   & \textbf{50.05}   \\
 & ConvNeXt‑Base \cite{liu2022convnext}  & Base & \underline{65.23} & \underline{96.18} & \underline{65.86} & \underline{65.73} & \underline{65.69} \\
 & & KAN  & \textbf{66.26}   & \textbf{96.54}   & \textbf{66.85}   & \textbf{66.62}   & \textbf{66.66}   \\
 & ViT‑Base \cite{dosovitskiy2020vit}    & Base & \underline{65.54} & \underline{96.43} & \underline{65.99} & \underline{66.15} & \underline{65.99} \\
 & & KAN  & \textbf{65.93}   & \textbf{96.57}   & \textbf{66.49}   & \textbf{66.55}   & \textbf{66.30}   \\
\midrule
\multirow{8}{*}{WikiArt}
  
 & EfficientNet‑B0 \cite{tan2019efficientnet} & Base & \textbf{50.09} & \underline{92.31} & \underline{50.81} & \textbf{49.63}   & \textbf{50.02}   \\
 & & KAN  & \underline{50.06} & \textbf{92.68}   & \textbf{51.23}   & \underline{48.91} & \underline{49.73} \\
 & ConvNeXt‑Base \cite{liu2022convnext}  & Base & \underline{60.08} & \underline{96.26} & \underline{61.37} & \underline{61.63} & \underline{61.46} \\
 & & KAN  & \textbf{60.95}   & \textbf{96.72}   & \textbf{62.00}   & \textbf{62.56}   & \textbf{62.22}   \\
 & ViT‑Base \cite{dosovitskiy2020vit}   & Base & \underline{61.75} & \underline{96.83} & \underline{64.97} & \underline{63.22} & \underline{63.44} \\
 & & KAN  & \textbf{61.98}   & \textbf{97.16}   & \textbf{65.93}   & \textbf{63.43}   & \textbf{64.04}   \\
\bottomrule
\end{tabular}
\end{adjustbox}
\end{table}

\section{Discussion}
\label{sec:Discussion}
\subsection{Effect of KAN Positioning Across Dual-Teacher Branches}
Table~\ref{tab:placement} examines the impact of placing the KAN projection head in different locations within the ConvNeXt-Base architecture on the Pandora18k dataset. Results indicate that inserting KAN into the student branch alone yields a 0.48\% increase in Top-1 accuracy, with additional improvements of 0.71\% in precision, 0.40\% in recall, and 0.44\% in F1-score, alongside a 0.10\% gain in Top-5 accuracy. This suggests KAN enhances the model's ability to disentangle complex nonlinear features while maintaining teacher signals and inserting KAN into the style teacher results in the highest Top-1 accuracy at 66.49\%, a 1.26\% improvement over the baseline, showcasing better texture and feature extraction. This placement also enhances recall and F1-score, increasing robustness to class imbalance. Conversely, placing KAN only in the momentum teacher yields minimal gains, indicating that its benefits are diminished outside the style-aware context. The most significant performance improvements occur when KAN is applied across all three components: the Student, the style teacher, and the momentum teacher, resulting in gains of 1.03\% in Top-1 accuracy, 0.36\% in Top-5 accuracy, 0.99\% in precision, 0.89\% in recall, and 0.97\% in F1-score. These findings highlight KAN’s effectiveness in integrating style-guided abstraction with temporal stability, improving feature clustering and inter-class separability in self-supervised learning.

\begin{table}[h]
\centering
\footnotesize
\setlength{\tabcolsep}{2pt}
\renewcommand{\arraystretch}{0.5}
\caption{Impact of KAN placement in ConvNeXt projection heads on \textbf{Pandora18k}.}
\label{tab:placement} 
\begin{adjustbox}{width=\linewidth}
\begin{tabular}{
  l 
  l 
  S[table-format=2.2] 
  S[table-format=2.2] 
  S[table-format=2.2] 
  S[table-format=2.2] 
  S[table-format=2.2] 
}
\toprule
\textbf{Dataset} & \textbf{Variant} & {\textbf{Top‑1\,(\%)}} & {\textbf{Top‑5\,(\%)}} & {\textbf{Prec.\,(\%)}} & {\textbf{Rec.\,(\%)}} & {\textbf{F1\,(\%)}} \\
\midrule
\multirow{5}{*}{Pandora18k}
 & Base (MLP)          & 65.23           & 96.18           & 65.86           & 65.73           & 65.69           \\
 & Student KAN         & 65.71           & \underline{96.28} & \underline{66.57} & \underline{66.13} & \underline{66.13} \\
 & Style Teacher KAN   & \textbf{66.49} & 96.05           & 66.49           & \underline{66.25} & 66.09           \\
 & Momentum Teacher KAN  & 65.47           & 96.21           & 65.90           & 65.80           & 65.85           \\
 & All Heads KAN (Ours)    & \underline{66.26} & \textbf{96.54} & \textbf{66.85} & \textbf{66.62} & \textbf{66.66} \\
\bottomrule
\end{tabular}
\end{adjustbox}
\end{table}

\subsection{Confusion Matrix Analysis}
\label{sec:confusion-matrix}
The confusion matrices for the Kernel-Adaptive Network (KAN) on the \textit{WikiArt} and \textit{Pandora18k} datasets reveal its strengths in structured classification as well as challenges with conceptual ambiguities. In the \textit{WikiArt} dataset, KAN achieves \textbf{83.2\%} accuracy for Northern Renaissance and \textbf{88.2\%} for Abstract Expressionism, though Baroque shows some misclassifications into Realism (\textbf{8.4\%}) and Romanticism (\textbf{6.9\%}). 
On the \textit{Pandora18k} dataset, KAN performs even better, with Abstract Expressionism at \textbf{96.4\%} and Baroque at \textbf{90.7\%}. Romanticism scores \textbf{77.1\%}, but Ukiyo-e is confused with Surrealism and Symbolism (\textbf{5.8\%} each), while Socialist Realism reaches \textbf{69.6\%} and overlaps with Realism (\textbf{7.0\%}). 
Cross-dataset comparisons show a reliance on low-level visual features, with Impressionism at \textbf{68.5\%} on \textit{WikiArt} and Magic Realism at only \textbf{53.1\%} on \textit{Pandora18k}, highlighting the challenges in classifying styles needing deeper cultural context.

\begin{figure}[t]
  \centering
  \begin{subfigure}[t]{0.48\linewidth}
    \centering
    \includegraphics[width=\linewidth]{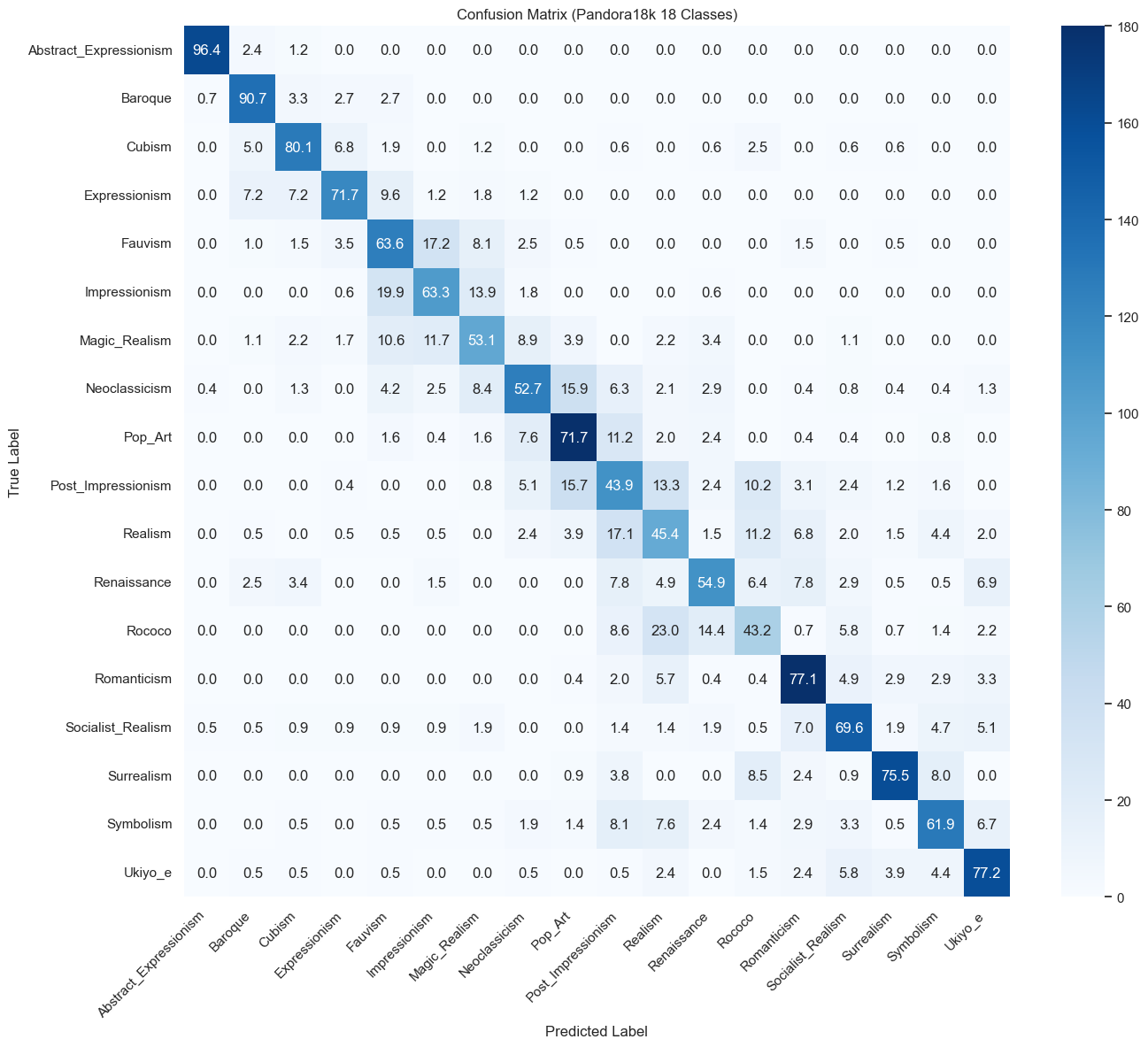}
    \caption{ConvNeXt-Base + KAN on Pandora18k. Darker diagonal entries indicate high true positive counts; sparse off-diagonals reveal low inter-class confusion.}
    \label{fig:conf_pandora}
  \end{subfigure}
  \hfill
  \begin{subfigure}[t]{0.48\linewidth}
    \centering
    \includegraphics[width=\linewidth]{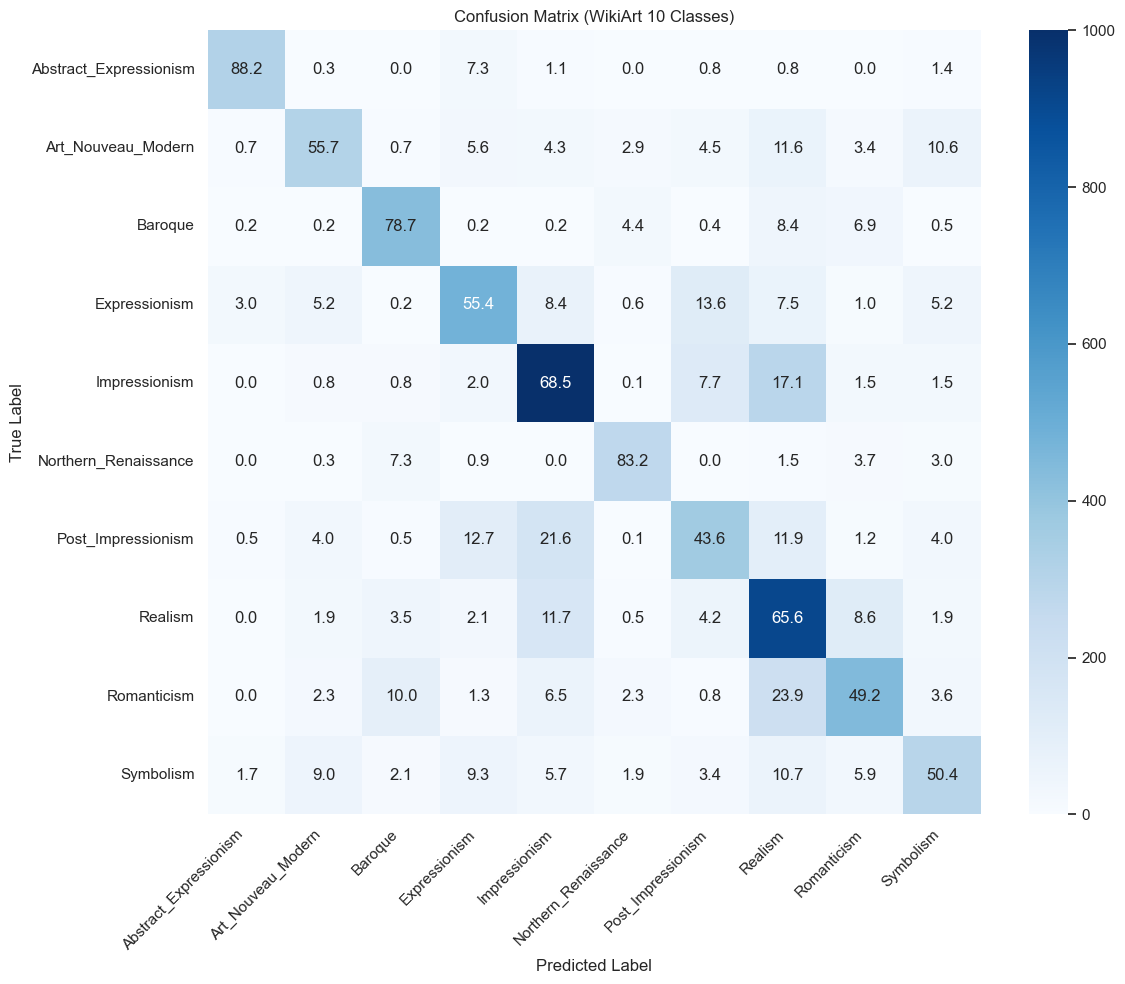}
    \caption{ConvNeXt-Base + KAN on WikiArt. Darker diagonal entries indicate high true positive counts; sparse off-diagonals reveal low inter-class confusion.}
    \label{fig:conf_wikiart}
  \end{subfigure}
  \caption{Class-wise confusion matrices for ConvNeXt-Base with KAN on (a) the Pandora18k and (b) WikiArt test sets. Strong diagonal dominance and minimal cross-style errors highlight the model’s robustness.}
  \label{fig:confusion_matrices_combined}
\end{figure}

\begin{figure}[h]
    \centering
    \begin{subfigure}[b]{0.48\textwidth}
        \includegraphics[width=\linewidth]{"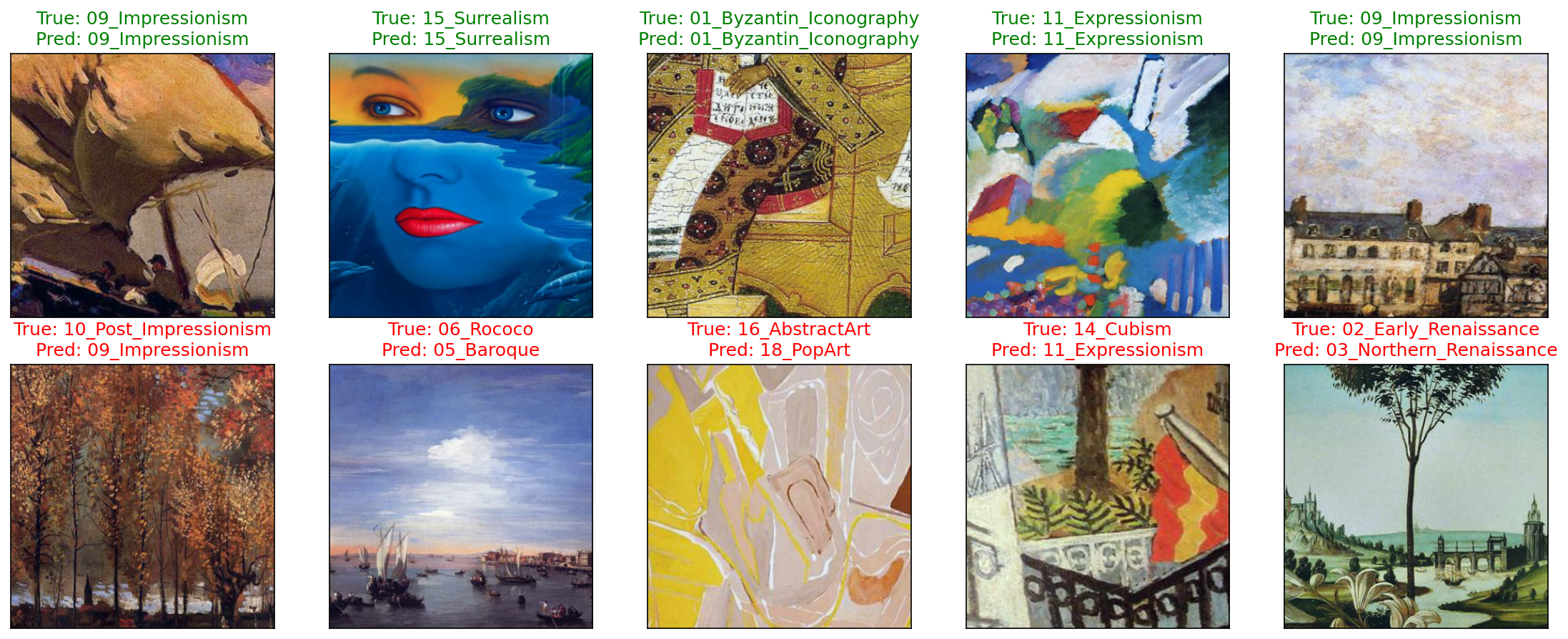"}
        \caption{Predictions from ConvNeXt-Base model}
        \label{fig:convbase_preds}
    \end{subfigure}
    \hfill
    \begin{subfigure}[b]{0.48\textwidth}
        \includegraphics[width=\linewidth]{"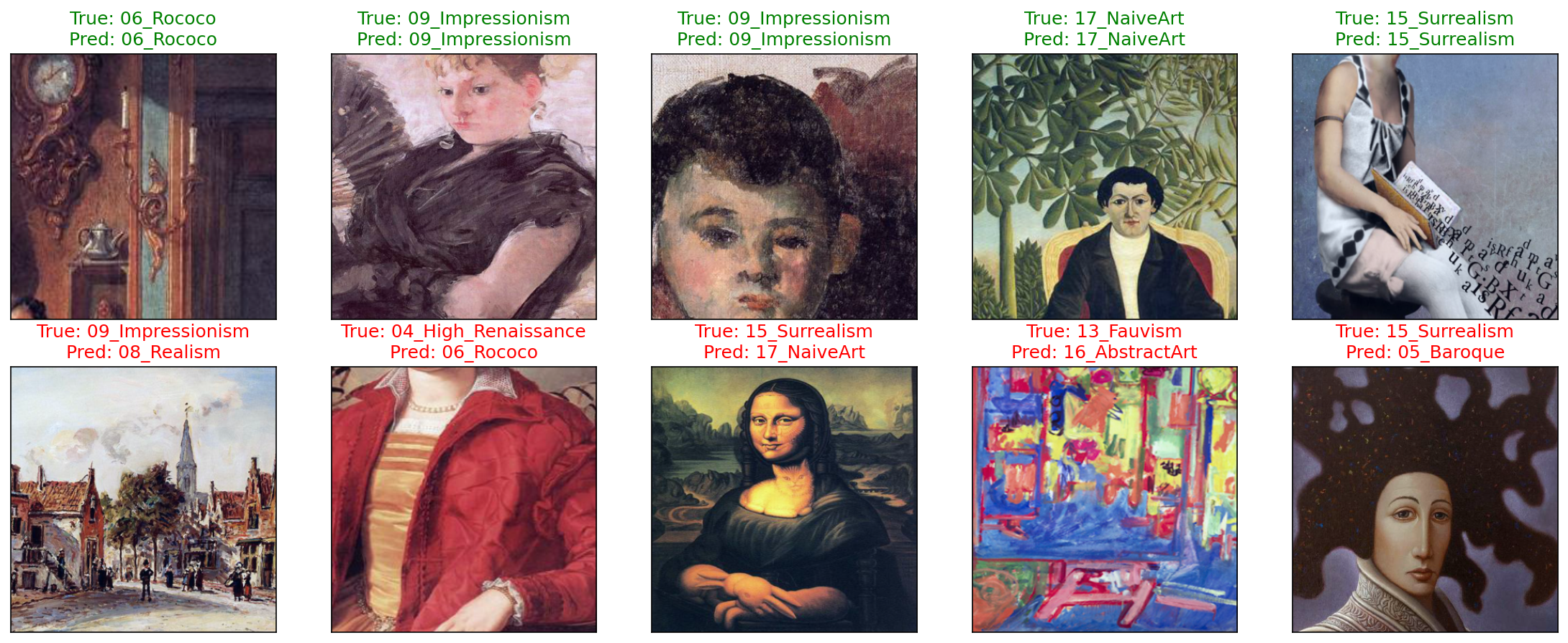"}
        \caption{Predictions from ConvNeXt-Base + KAN model}
        \label{fig:resnet_preds}
    \end{subfigure}
    \caption{Comparative prediction examples from different models on the Pandora18k dataset. (a) Results from the convolutional baseline model, (b) Results from the ConvNeXt-Base architecture with KAN.}
    \label{fig:model_comparison}
\end{figure}

\section{Conclusion}
\label{sec:conclusion}
This study introduces a self-supervised methodology for art style analysis that integrates Kolmogorov–Arnold Networks (KANs) with Gramian style alignment. By replacing traditional multi-layer perceptron (MLP) projection heads with spline-activated KAN layers, we address limitations of linear feature interactions while ensuring computational efficiency often lost in multi-teacher paradigms. Our framework significantly improves cross-dataset generalization across established benchmarks like WikiArt and Pandora18k, outperforming previous dual-teacher approaches through adaptive nonlinear projection~\cite{luo2025}.
Analysis of prediction patterns (\Cref{fig:model_comparison}) reveals challenges in distinguishing stylistically adjacent movements, particularly between High Renaissance/Rococo (\Cref{fig:convbase_preds}) and Post-Impressionism/Expressionism (\Cref{fig:resnet_preds}). These misclassifications suggest that finer aspects of painterly technique might require geometric disentanglement beyond current alignment methods.


%
%
\bibliographystyle{splncs04} 
\bibliography{main}

\begin{thebibliography}{10}
\providecommand{\url}[1]{\texttt{#1}}
\providecommand{\urlprefix}{URL }
\providecommand{\doi}[1]{https://doi.org/#1}

\bibitem{cetinic2018fine}
Cetinic, E., Lajic, T., Grgic, S.: Fine-tuning convolutional neural networks for fine art classification. Expert Systems with Applications pp. 107--118 (2018)

\bibitem{cetinic2019}
Cetinic, E., Lipic, T., Grgic, S.: Towards interpreting deep learning-based art style classification. arXiv preprint arXiv:1908.04307  (2019)

\bibitem{chen2020simclr}
Chen, T., Kornblith, S., Norouzi, M., Hinton, G.: A simple framework for contrastive learning of visual representations pp. 1597--1607 (2020)

\bibitem{chen2020}
Chen, T., Kornblith, S., Norouzi, M., Hinton, G.: A simple framework for contrastive learning of visual representations. In: ICML. pp. 1597--1607 (2020)

\bibitem{kan2024nonlinear}
Doe, A., Nguyen, L., Patel, R.: Kolmogorov--arnold networks: Learning nonlinear projection through spline-based activation. In: Proceedings of the International Conference on Machine Learning. pp. 1023--1032 (2024)

\bibitem{ressl2023relation}
Doe, J., Smith, J.: Relational consistency via gram matrix alignment for self-supervised learning. In: IEEE Transactions on Pattern Analysis and Machine Intelligence. vol.~45, pp. 234--245 (2023)

\bibitem{dosovitskiy2020vit}
Dosovitskiy, A., Beyer, L., Kolesnikov, A., Weissenborn, D., Zhai, X., et~al.: An image is worth 16x16 words: Transformers for image recognition at scale. arXiv preprint arXiv:2010.11929  (2020)

\bibitem{elgammal2018}
Elgammal, A., Liu, B., Elhoseiny, M., Mazzone, M.: The shape of art history in the eyes of the machine. In: AAAI Conference on Artificial Intelligence (2018)

\bibitem{falomir2018qartlearn}
Falomir, Z., Museros, L., Sanz, I., Gonzalez-Abril, L.: Categorizing paintings in art styles based on qualitative color descriptors, quantitative global features and machine learning (qart-learn). Expert Systems with Applications pp. 83--94 (2018)

\bibitem{7926652}
Florea, C., Toca, C., Gieseke, F.: Artistic movement recognition by boosted fusion of color structure and topographic description. In: 2017 IEEE Winter Conference on Applications of Computer Vision (WACV). pp. 569--577 (2017). \doi{10.1109/WACV.2017.69}

\bibitem{fukuda2017}
Fukuda, T., Suzuki, M., Kurata, G., Thomas, S., Cui, J., Ramabhadran, B.: Efficient knowledge distillation from an ensemble of teachers. In: Interspeech. pp. 3697--3701 (2017)

\bibitem{fukuda2017efficient}
Fukuda, T., Suzuki, M., Kurata, G., Thomas, S., Cui, J., Ramabhadran, B.: Efficient knowledge distillation from an ensemble of teachers. In: Interspeech. pp. 3697--3701 (2017)

\bibitem{gairola2020unsupervised}
Gairola, S., Shah, R., Narayanan, P.: Unsupervised image style embeddings for retrieval and recognition tasks. In: IEEE/CVF Winter Conference on Applications of Computer Vision. pp. 2381--2399 (2020)

\bibitem{garg2022selflabelingrefinementrobustrepresentation}
Garg, S., Jain, D.: Self-labeling refinement for robust representation learning with bootstrap your own latent (2022), \url{https://arxiv.org/abs/2204.04545}

\bibitem{he2020moco}
He, K., Fan, H., Wu, Y., Xie, S., Girshick, R.: Momentum contrast for unsupervised visual representation learning pp. 9729--9738 (2020)

\bibitem{he2020}
He, K., Fan, H., Wu, Y., Xie, S., Girshick, R.: Momentum contrast for unsupervised visual representation learning. In: CVPR. pp. 9729--9738 (2020)

\bibitem{hinton2015distilling}
Hinton, G., Vinyals, O., Dean, J.: Distilling the knowledge in a neural network. In: Neural Information Processing Systems Deep Learning and Representation Learning Workshop (2015)

\bibitem{johnson2017neural}
Johnson, J.: Neural style representations and the large-scale classification of artistic style pp. 283--285 (2017)

\bibitem{karayev2013recognizing}
Karayev, S., Hertzmann, A., Winnemoeller, H., Agarwala, A., Darrell, T.: Recognizing image style. In: British Machine Vision Conference (2013)

\bibitem{Karayev_2014_BMVC}
Karayev, S., Trentacoste, M., Han, H., Agarwala, A., Darrell, T., Hertzmann, A., Winnemoeller, H.: Recognizing image style. In: British Machine Vision Conference (BMVC). Nottingham, UK (September 2014), \url{https://bmva-archive.org.uk/bmvc/2014/files/paper121/index.html}

\bibitem{karayev2013stylearxiv}
Karayev, S., et~al.: Recognizing image style. arXiv preprint arXiv:1311.3715  (2013)

\bibitem{krizhevsky2012imagenet}
Krizhevsky, A., Sutskever, I., Hinton, G.E.: Imagenet classification with deep convolutional neural networks. In: Advances in Neural Information Processing Systems. pp. 1097--1105 (2012)

\bibitem{liu2022convnext}
Liu, Z., Mao, H., Wu, C.Y., Feichtenhofer, C., Darrell, T., Xie, S.: A convnet for the 2020s. arXiv preprint arXiv:2201.03545  (2022)

\bibitem{liu2024kan}
Liu, Z., Meng, Y., Wang, Y., Zhang, G.: Kan: Kolmogorov-arnold networks. arXiv preprint arXiv:2404.19756  (2024)

\bibitem{liu2024}
Liu, Z., Wang, Y., Vaidya, S., Ruehle, F., Halverson, J., Solja{\v{c}}i{\'c}, M., Hou, T.Y., Tegmark, M.: Kan: Kolmogorov-arnold networks. arXiv preprint arXiv:2404.19756  (2024)

\bibitem{luo2025}
Luo, M., Liu, L., Lu, Y., Suen, C.Y.: Art style classification via self-supervised dual-teacher knowledge distillation. Applied Soft Computing  \textbf{174},  112964 (2025)

\bibitem{luo2025dual}
Luo, M., Liu, L., Lu, Y., Suen, C.Y.: Art style classification via self-supervised dual-teacher knowledge distillation. Applied Soft Computing  \textbf{174},  112964 (2025)

\bibitem{peterson2018learninghierarchicalvisualrepresentations}
Peterson, J.C., Soulos, P., Nematzadeh, A., Griffiths, T.L.: Learning hierarchical visual representations in deep neural networks using hierarchical linguistic labels (2018), \url{https://arxiv.org/abs/1805.07647}

\bibitem{pham2023collaborative}
Pham, C., Hoang, T., Do, T.T.: Collaborative multi-teacher knowledge distillation for learning low bit-width deep neural networks. In: IEEE/CVF Winter Conference on Applications of Computer Vision (WACV). pp. 6435--6443 (2023)

\bibitem{pham2020distillation}
Pham, V., et~al.: Multi-teacher multi-task knowledge distillation for semantic segmentation. In: IEEE/CVF Conference on Computer Vision and Pattern Recognition (2020)

\bibitem{simonyan2014very}
Simonyan, K., Zisserman, A.: Very deep convolutional networks for large-scale image recognition. In: International Conference on Learning Representations (2015)

\bibitem{tan2019efficientnet}
Tan, M., Le, Q.V.: Efficientnet: Rethinking model scaling for convolutional neural networks  (2019)

\bibitem{zhang2023multitask}
Zhang, W., Li, M., Chen, J.: A multi-task self-supervised framework for art style classification. In: Proceedings of the IEEE International Conference on Computer Vision. pp. 1234--1242 (2023)

\bibitem{zhang2022threeelement}
Zhang, Y., Chen, H., Wang, L., Yang, M.H., Li, W.: Three-element learning: A unified framework for multi-task self-supervised representation learning. In: Proceedings of the IEEE/CVF Conference on Computer Vision and Pattern Recognition (CVPR). pp. 11234--11244 (2022)

\bibitem{zhou2024three}
Zhou, D., Zhou, D., Wei, G., Yuan, X.: Three-dimensional shape reconstruction from digital freehand design sketching based on deep learning techniques. Applied Sciences  \textbf{14}(24),  11717 (2024). \doi{10.3390/app142411717}

\end{thebibliography}
\end{document}